\UseRawInputEncoding

\documentclass[10pt,twocolumn,letterpaper]{article}

\usepackage[pagenumbers]{cvpr} % To force page numbers, e.g. for an arXiv version

\usepackage[accsupp]{axessibility}
\usepackage{graphicx}
\usepackage{amsmath}
\usepackage{amssymb}
\usepackage{booktabs}
\usepackage{tabularx,colortbl}
\usepackage{multirow}
\usepackage{makecell}
\usepackage{capt-of,etoolbox}
\usepackage[american]{babel}
\usepackage[table]{xcolor}

\usepackage[pagebackref,breaklinks,colorlinks]{hyperref}

\usepackage[capitalize]{cleveref}
\crefname{section}{Sec.}{Secs.}
\Crefname{section}{Section}{Sections}
\Crefname{table}{Table}{Tables}
\crefname{table}{Tab.}{Tabs.}

\def\cvprPaperID{3567} % *** Enter the CVPR Paper ID here
\def\confName{CVPR}
\def\confYear{2023}

\begin{document}

\title{A Unified Pyramid Recurrent Network for Video Frame Interpolation}

\author{Xin Jin$^1$ ~~~Longhai Wu$^1$ ~~Jie Chen$^1$ ~~~Youxin Chen$^1$ 
    ~~~Jayoon Koo$^2$ ~~~Cheul-hee Hahm$^2$ \\ $^1$Samsung Electronics
    (China) R\&D Center ~~~~~~~~~$^2$Samsung Electronics, South Korea \\
    {\tt\small
    \{xin.jin,longhai.wu,ada.chen,yx113.chen,j.goo,chhahm\}@samsung.com}
}

% \and

\maketitle

%%%%%%%%% ABSTRACT
\begin{abstract}

    Flow-guided synthesis provides a common framework for frame interpolation,
    where optical flow is estimated to guide the synthesis of intermediate
    frames between consecutive inputs. In this paper, we present UPR-Net, a
    novel Unified Pyramid Recurrent Network for frame interpolation. Cast in a
    flexible pyramid framework, UPR-Net exploits lightweight recurrent modules
    for both bi-directional flow estimation and intermediate frame synthesis. At
    each pyramid level, it leverages estimated bi-directional flow to generate
    forward-warped representations for frame synthesis; across pyramid levels,
    it enables iterative refinement for both optical flow and intermediate
    frame. In particular, we show that our iterative synthesis strategy can
    significantly improve the robustness of frame interpolation on large motion
    cases. Despite being extremely lightweight (1.7M parameters), our base
    version of UPR-Net achieves excellent performance on a large range of
    benchmarks. Code and trained models of our UPR-Net series are available at:
    \url{https://github.com/srcn-ivl/UPR-Net}.

\end{abstract}

%%%%%%%%% BODY TEXT
\section{Introduction}
\label{sec:intro}

Video frame interpolation (VFI) is a classic low-level vision task. It aims to
increase the frame rate of videos, by synthesizing non-existent intermediate
frames between consecutive frames. VFI technique supports many practical
applications including novel view synthesis~\cite{flynn2016deepstereo}, video
compression~\cite{lu2017novel}, cartoon creation~\cite{siyao2021deep}, \etc.

Despite great potential in applications, video frame interpolation remains an
unsolved problem, due to challenges like complex and large motions, occlusions,
and illumination changes in real-world videos. Depending on whether or not
optical flow is incorporated to compensate for inter-frame motion, existing
methods can be roughly classified into two categories: flow-agnostic
methods~\cite{niklaus2017video,meyer2018phasenet,cheng2020video,choi2020channel},
and flow-guided
synthesis~\cite{jiang2018super,liu2017video,niklaus2018context,bao2019memc,niklaus2020softmax,park2020bmbc,park2021asymmetric}.
With recent advances in optical
flow~\cite{ilg2017flownet,hui2018liteflownet,sun2018pwc,teed2020raft},
flow-guided synthesis has developed into a popular framework with compelling
performance for video frame interpolation.

\begin{figure}[tb]
\centering
\includegraphics[width=0.48\textwidth]{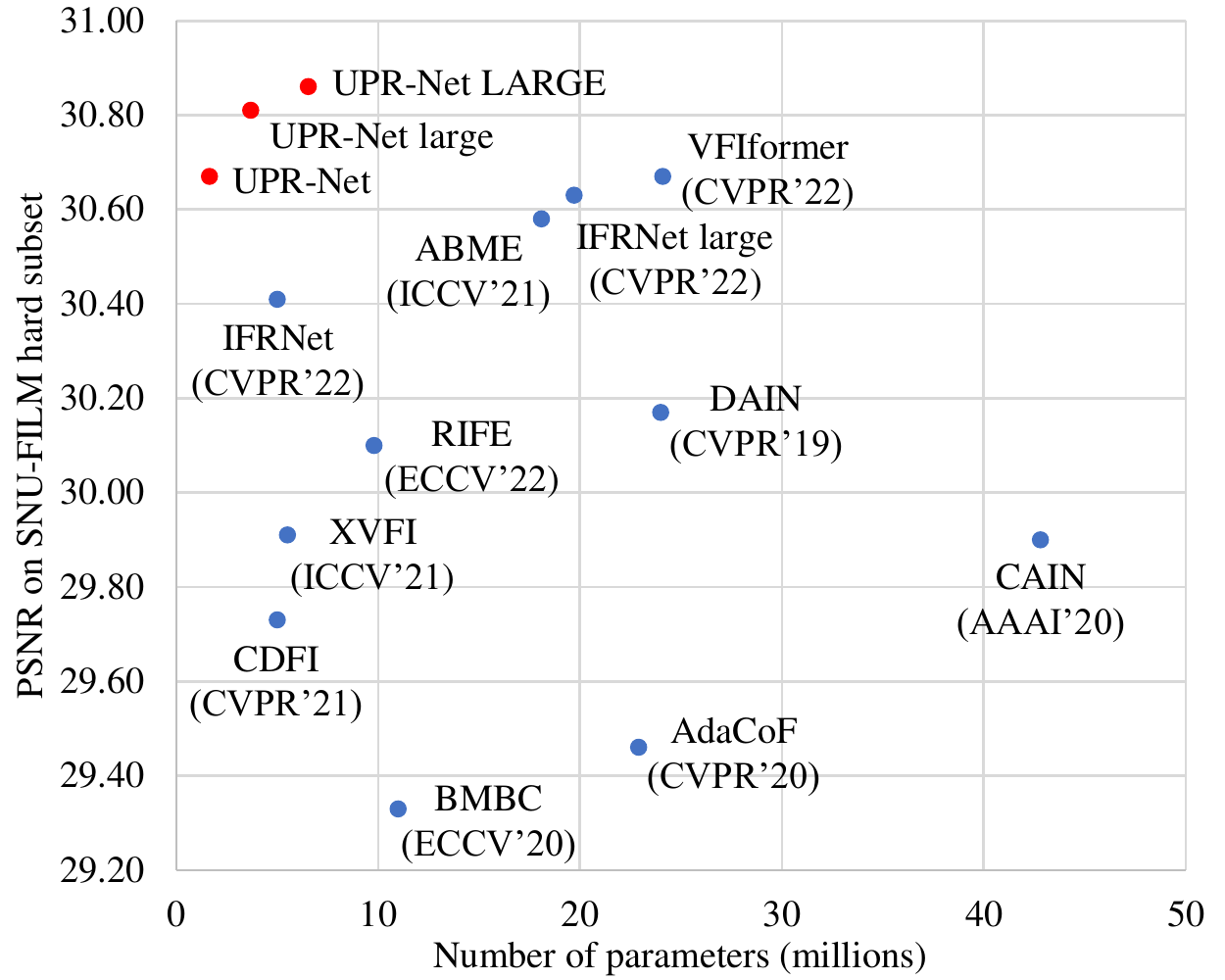}
\caption{Comparison of performance and model size on the hard subset of SNU-FILM
benchmark~\cite{choi2020channel}.  Our UPR-Net series achieve state-of-the-art
accuracy with extremely small parameters.}
\vspace{-0.25cm}
\label{fig:intro}
\end{figure}

Most of existing flow-guided methods follow a similar procedure: estimating
optical flow for desired time step, warping input frames and their context
features based on optical flow, and synthesizing intermediate frame from warped
representations. Where technical choices may diverge in this procedure, is the
warping operation and the optical flow it requires.  Backward-warping is
traditionally used for frame
interpolation~\cite{jiang2018super,liu2017video,bao2019memc,park2020bmbc,park2021asymmetric},
but acquiring high-quality bilateral intermediate flow for it is often
challenging. Forward-warping can directly use linearly-scaled bi-directional
flow between input frames (which is easier to obtain), and thus has recently
emerged as a promising direction for frame
interpolation~\cite{niklaus2018context,niklaus2020softmax}.

% For forward-warping, ambiguities where multiple source pixels are mapped to same
% target should be addressed~\cite{niklaus2020softmax}. 

In common flow-guided synthesis
pipeline~\cite{bao2019depth,niklaus2020softmax,park2021asymmetric,lee2022enhanced},
optical flow is typically estimated from coarse to fine by a pyramid network,
but intermediate frame is synthesized just once by a synthesis network. Despite
promising performance on low-resolution videos, this practice misses the
opportunity of iteratively refining the interpolation for high-resolution
inputs. Second, for large motion cases, an important issue has been overlooked
by previous works: even when estimated motion is visually plausible, in many
cases, the obvious artifacts in warped frames (\eg, large holes in
forward-warped frames) may also degrade the interpolation performance.  Last,
existing methods typically rely on heavy model architectures to achieve good
performance, blocking them from being deployed on platforms with limited
resources, \eg, mobile devices.

Aiming at these issues, we introduce UPR-Net, a novel Unified Pyramid Recurrent
Network for frame interpolation. Within a pyramid framework, UPR-Net exploits
lightweight recurrent modules for both bi-directional flow estimation and
forward-warping based frame synthesis. It enables iterative refinement of both
optical flow and intermediate frame across pyramid levels, producing compelling
results on complex and large motion cases.

Our work draws inspirations from many existing works, but is significantly
distinguished from them in three aspects. First, UPR-Net inherits the merit of
recent pyramid recurrent bi-directional flow
estimators~\cite{sim2021xvfi,jin2022enhanced}, allowing to customize the number
of pyramid levels in testing to estimate extremely large motions. But, it goes
one step further, by exploiting pyramid recurrent network for coarse-to-fine
frame synthesis, and unifying motion estimation and frame synthesis within a
single pyramid recurrent network.

Second, we reveal that our coarse-to-fine iterative synthesis can significantly
improve the robustness of frame interpolation on large motion cases. At
high-resolution pyramid levels, forward-warped frames may suffer from obvious
holes due to large motions, resulting in poor interpolation for many cases. We
show that this issue can be remedied to a large extent, by feeding the frame
synthesis module with the intermediate frame estimate upsampled from previous
lower-resolution pyramid level.

Third, both of our optical flow and frame synthesis modules are extremely
lightweight. Yet, they are still carefully integrated with the key ingredients
from modern researches on optical flow~\cite{sun2018pwc,teed2020raft} and frame
synthesis~\cite{niklaus2018context}. Specifically, at each pyramid level,
UPR-Net firstly extracts CNN features for input frames, then constructs a
correlation volume for simultaneous bi-directional flow estimation. It predicts
refined intermediate frame from forward-warped input frames and their CNN
features, along with upsampled intermediate frame estimate.

We conduct extensive experiments to verify the effectiveness of UPR-Net for
frame interpolation. Our base version of UPR-Net only has 1.7M parameters. Yet,
it achieves excellent performance on both low- and high-resolution
benchmarks, when trained with low-resolution data. Figure~\ref{fig:intro} gives
a comparison of accuracy and model size on the hard subset of
SNU-FILM~\cite{choi2020channel}, where our UPR-Net series achieve
state-of-the-art accuracy with much fewer paprameters. In addition, we validate
various design choices of UPR-Net by ablation studies.

\section{Related Work}
\label{sec:related}

\paragraph{Pyramid recurrent optical flow estimator.} PWC-Net~\cite{sun2018pwc}
has been traditionally used for optical flow by frame interpolation
methods~\cite{niklaus2018context,bao2019depth,niklaus2020softmax}.  However, the
fixed number of pyramid levels makes it difficult to handle extremely large
motions beyond the training phase. Recently, pyramid recurrent optical flow
estimators~\cite{zhang2020flexible,sim2021xvfi} are developed to handle large
motion, by sharing the structure across pyramid levels and customizing the
pyramid levels in testing. A larger number of pyramid levels can better handle
large motions that often appear in high-resolution videos.

Previous pyramid recurrent estimators typically employ a plain U-Net as the base
estimator at each pyramid level. However, U-Net is over-simplified for optical
flow due to the lack of correlation volume~\cite{sun2018pwc,teed2020raft}.  Very
recently, EBME~\cite{jin2022enhanced} incorporates correlation volume into
pyramid recurrent network for simultaneous bi-directional flow estimation.  We
follow the basic idea in~\cite{jin2022enhanced} for bi-directional flow, but
modify it to better adapt to our unified pyramid network for frame
interpolation.

\paragraph{Coarse-to-fine image synthesis.}  Coarse-to-fine processing is a
mature technology for high-resolution image synthesis, where low-resolution
images are firstly synthesized, and then iteratively refined until generating
the desired high-resolution output. It has many successful applications,
including photographic image synthesis conditioned on semantic
layouts~\cite{chen2017photographic}, adversarial image
generation~\cite{denton2015deep}, and recent diffusion model based image
synthesis~\cite{esser2021imagebart}.

However, coarse-to-fine synthesis has been largely overlooked by existing frame
interpolation methods. Zhang \etal~\cite{zhang2020flexible} iteratively estimate
the occlusion mask within a pyramid recurrent framework, but still needs an
extra refinement network to obtain the final result. XVFI~\cite{sim2021xvfi}
estimates multi-scale intermediate frames during training, but does not perform
iterative refinement of intermediate frame.  IFRNet~\cite{kong2022ifrnet}
gradually refines the intermediate feature (rather than frame) until generating
the desired output, but it is not recurrent, and has limited capacity in
handling large motion. In this work, we iteratively refine the intermediate
frame within a pyramid recurrent framework.

\paragraph{Artifacts in warped frames.}
Although warping can compensate for per-pixel motion, it often creates
distortion and artifacts.  If a pixel is moved to a new location, and no other
pixels are moved to fill the old location, this pixel will appear twice in
backward-warped frame~\cite{lu2020devon,lee2022enhanced}, or leave a hole at
original location in forward-warped frame~\cite{niklaus2018context}.

\begin{figure*}[!htb]
\centering
\includegraphics[width=0.85\textwidth]{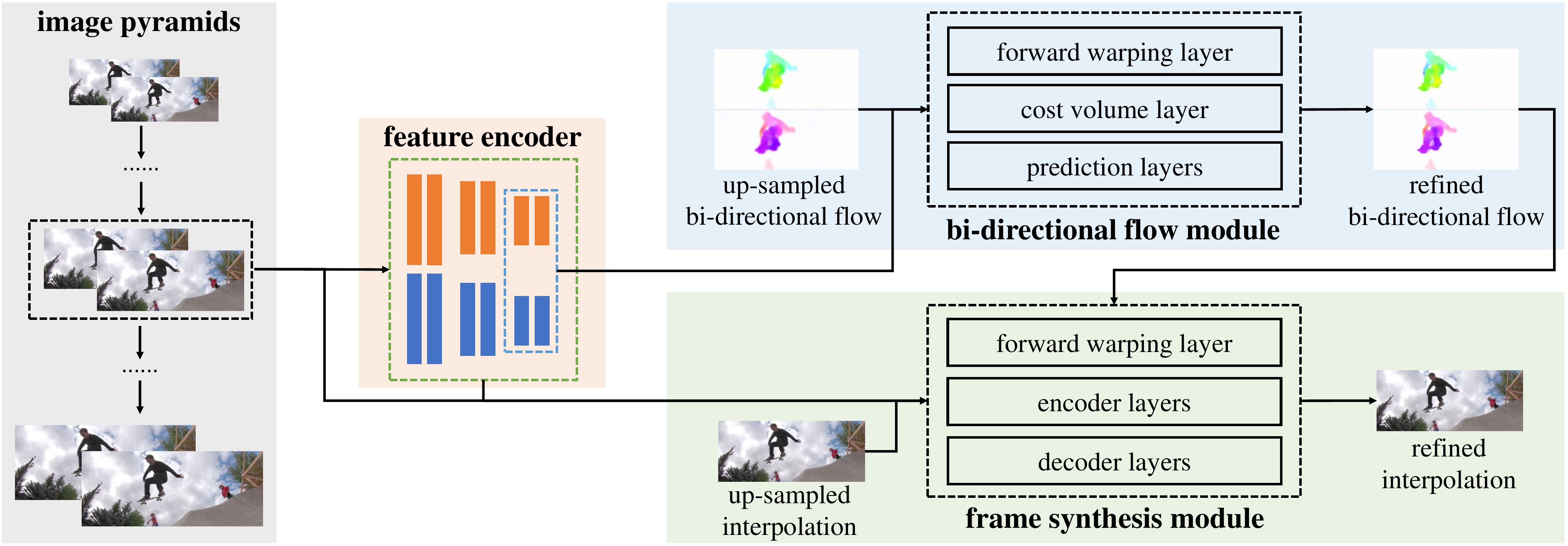}
\caption{Overview of our UPR-Net. Given two input frames, we first construct
    image pyramids for them, then apply a \textbf{recurrent} structure across
    pyramid levels to repeatedly refine estimated bi-directional flow and
    intermediate frame. Our recurrent structure consists of a feature encoder
    that extracts multi-scale features for input frames, a bi-directional flow
    module that refines bi-directional flow with correlation-injected features,
    and a frame synthesis module that refines intermediate frame estimate with
    forward-warped representations.
}
\vspace{-0.25cm}
\label{fig:pipeline}
\end{figure*}

To robustly synthesize intermediate frame from warped frames, existing frame
interpolation methods typically feed the synthesis network with both warped
frames and their context
features~\cite{niklaus2018context,niklaus2020softmax,huang2020rife}.  The
synthesis network can leverage rich contextual cues to infer the intermediate
frame from warped representations, even when artifacts exist in warped frames.

In this work, we observe that the synthesis network does work well for small
motion cases where forward-warped frames contain slight artifacts.
However, in presence of large motions, in many cases, the obvious holes in
forward-warped frames may lead to artifacts in interpolation.  We show that
our iterative synthesis can significantly improve the robustness of frame
interpolation on large motion cases.

Additionally, during forward-warping, the conflicted pixels mapped to the same
target should be addressed by simple averaging or certain weighted averaging
operation (\eg, softmax splatting~\cite{niklaus2020softmax}).  In this work, we
adopt the average splatting~\cite{niklaus2020softmax} as forward-warping for
simplicity.

\section{Our Approach}
\label{sec:approach}

% In this section, we firstly overview our UPR-Net for frame interpolation. Then,
% we detail the designs of three key modules of UPR-Net, which are shared across
% pyramid levels.  Lastly, we describe resolution-ware adaptation in testing,
% and architecture variants of UPR-Net.

\subsection{Unified Pyramid Recurrent Network}

We illustrate the overall pipeline of UPR-Net in Figure~\ref{fig:pipeline}. It
unifies bi-directional flow estimation and frame synthesis within a pyramid
structure, and shares the weights across pyramid levels. This macro pyramid
recurrent architecture has two advantages: (\romannumeral1) reducing the
parameters of the full pipeline; (\romannumeral2) allowing to customize the
number of pyramid levels in testing to handle large motions.

Given a pair of consecutive frames $I_0$, $I_1$, and the desired time step $t$
($0 \leq t \leq 1$), our goal is to synthesize the non-existent intermediate
frame $I_t$.  UPR-Net tackles this task via an iterative refinement procedure
across $L$ image pyramid levels, from the top level with down-sampled frames
$I_0^{L-1}$ and $I_1^{L-1}$, to the bottom (zeroth) pyramid level with the
original input frames $I_0^0$ and $I_1^0$.

At each pyramid level, UPR-Net employs a feature encoder to extract multi-scale
CNN features for both input frames. Then, the features from the last layer of
feature encoder and the optical flow upsampled from previous level are processed
by a bi-directional flow module to produce refined bi-directional flow. The
refined optical flow is leveraged to forward-warp input frames and multi-scale
CNN features.  Combining warped representations with the interpolation upsampled
from previous level, a frame synthesis module is employed to generate refined
intermediate frame. This estimation process is repeated until generating the
final interpolation at the bottom pyramid level.

\subsection{Recurrent Frame Interpolation Modules}
Our recurrent structure consists of three lightweight modules: feature encoder,
bi-directional flow module, and frame synthesis module.

\paragraph{Feature encoder.} For optical flow~\cite{sun2018pwc,teed2020raft}, a
common practice is to extract per-pixel features and construct correlation
volume with features. For frame interpolation, it is also common to feed the
synthesis network with warped features to provide contextual
cues~\cite{niklaus2020softmax,huang2020rife}.  Considering these, at each
pyramid level, we firstly employ a feature encoder to extract multi-scale
features for input frames.

Our feature encoder has three convolutional stages: stage-0, stage-1, and
stage-2. Each stage consists of four convolutional layers, and the first layers
of the stage-1 and stage-2 perform down-sampling. We use the features from each
stage's last convolutional layer. Given images $I_0^l$ and $I_1^l$ at the $l$-th
pyramid level, we denote the features as $\{C_0^{l,0}, C_1^{l,0}\}$,
$\{C_0^{l,1}, C_1^{l,1}\}$ and $\{C_0^{l,2}, C_1^{l,2}\}$ for $I_0$ and $I_1$
extracted at stage 0, 1, 2, respectively. $\{C_0^{l,2}, C_1^{l,2}\}$ will be
used for bi-directional flow estimation, and all multi-scale features will be
used for context-aware frame synthesis.

\paragraph{Bi-directional flow module.}

\begin{figure}[tb]
\centering
\includegraphics[width=0.48\textwidth]{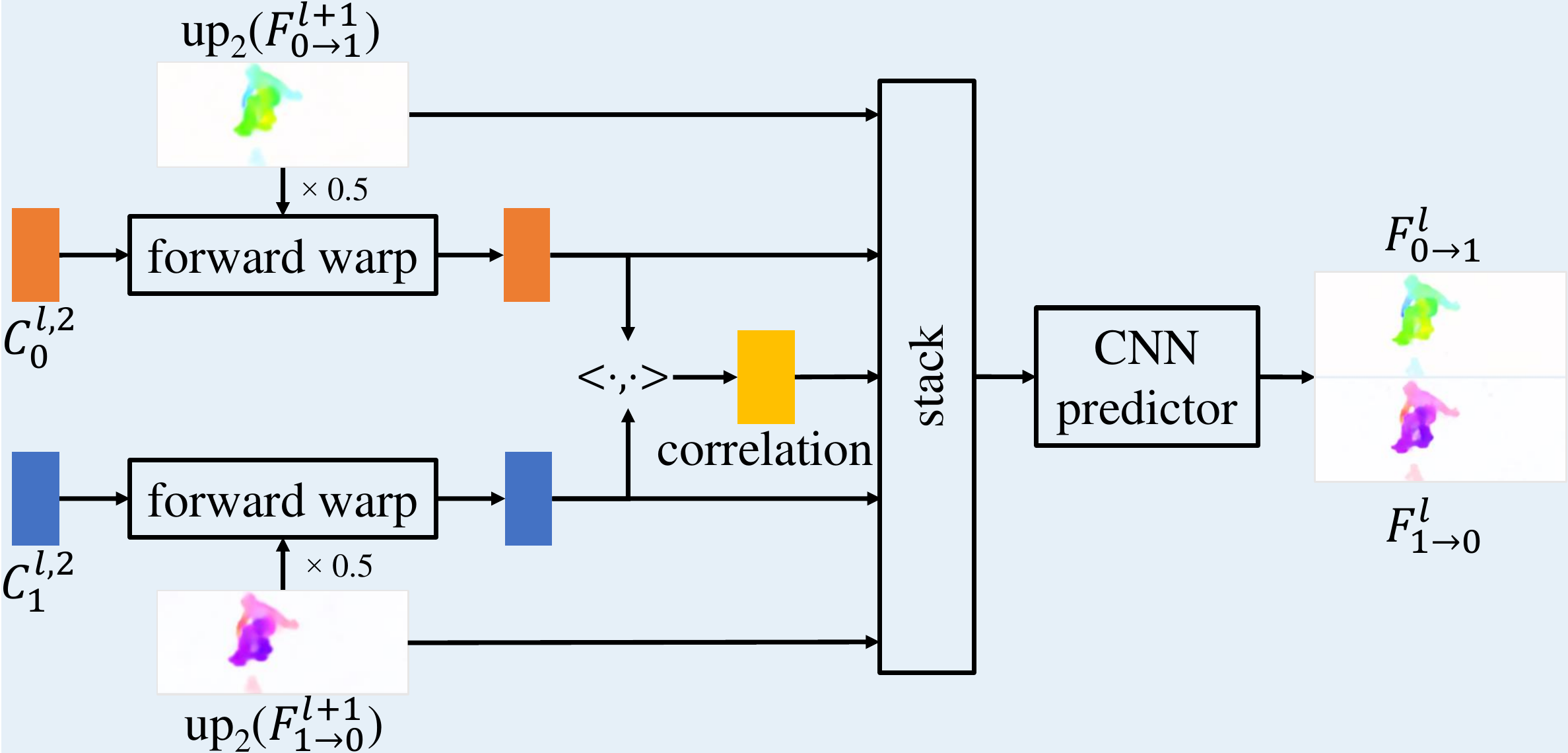}
\caption{Our bi-directional flow module.}
\vspace{-0.2cm}
\label{fig:flow-module}
\end{figure}

Let $F_{0 \rightarrow 1}^{l+1}$ and $F_{1 \rightarrow 0}^{l+1}$ denote the
refined bi-directional optical flow at level $l+1$.  At the $l$-th image pyramid
level, we firstly initialize the bi-directional flow by $\times$2 upsampling
$F_{0 \rightarrow 1}^{l+1}$ and $F_{1 \rightarrow 0}^{l+1}$: $ \hat{F}_{0
\rightarrow 1}^l = \mathrm{up}_2(F_{0 \rightarrow 1}^{l+1})$, $\hat{F}_{1
\rightarrow 0}^l = \mathrm{up}_2(F_{1 \rightarrow 0}^{l+1})$.  In particular,
the initial flow at top level is set to zero.  Based on initial flow, we can
obtain the optical flow from the input frames $I_0^l$ and $I_1^l$ to the hidden
middle frame $I_{0.5}^l$ by linear scaling: $ \hat{F}_{0 \rightarrow 0.5}^l =
0.5 \cdot \hat{F}_{0 \rightarrow 1}^l$, $\hat{F}_{1 \rightarrow 0.5}^l = 0.5
\cdot \hat{F}_{1 \rightarrow 0}^l$.

With $\hat{F}_{0 \rightarrow 0.5}^l$ and $\hat{F}_{1 \rightarrow 0.5}^l$, we
forward-warp the CNN features $C_0^{l,2}$ and $C_1^{l,2}$ to the middle frame to
align their pixels. Then, we construct a partial correlation
volume~\cite{sun2018pwc} using warped features,
and use a 6-layer CNN to predict refined bi-directional flow $F_{0
\rightarrow 1}^l$ and $F_{1 \rightarrow 0}^l$.  In particular, the input to the
CNN predictor is the concatenation of correlation volume, warped features,
initial flow $\hat{F}_{0 \rightarrow 1}^l$ and $\hat{F}_{1 \rightarrow 0}^l$,
and the upsampled feature from the 5-th layer of the CNN predictor at previous
pyramid level. Since the warped features are of 1/4 resolution of input frames,
the predicted optical flow is also of 1/4 resolution.  We use bi-linear
interpolation to up-sample optical flow to original scale.

Figure~\ref{fig:flow-module} illustrates our bi-directional flow module. Its
high-level design is similar to EBME~\cite{jin2022enhanced}. We adapt EBME for
iterative synthesis, by using the features from feature encoder shared with
synthesis module, and forward-warping CNN features rather than input frames.

\paragraph{Frame synthesis module.}

\begin{figure}[tb]
\centering
\includegraphics[width=0.48\textwidth]{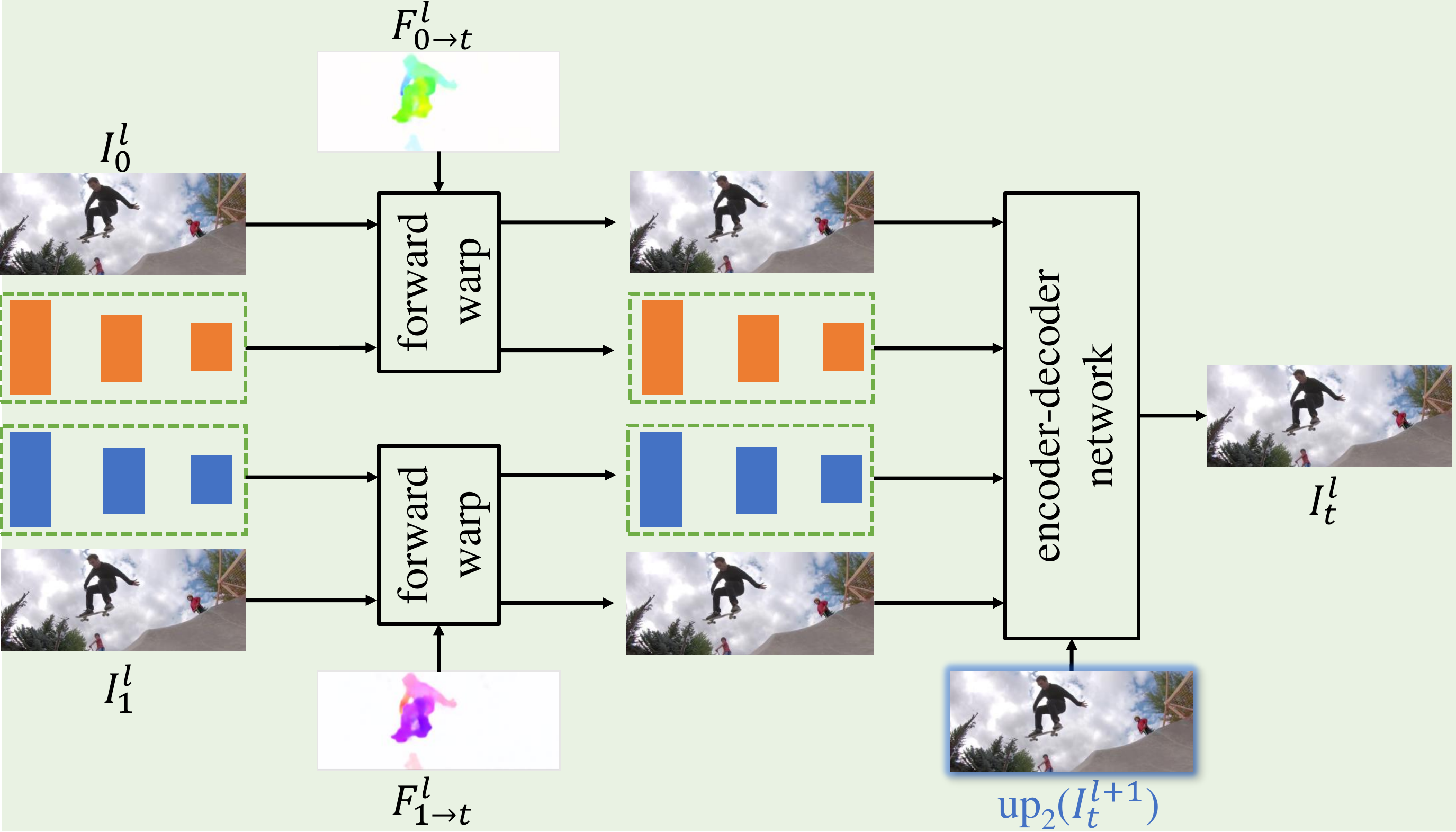}
\caption{Our frame synthesis module.}
\vspace{-0.25cm}
\label{fig:synthesis-module}
\end{figure}

Our frame synthesis module is based on a U-Net structure. The encoder part has
three convolutional stages, and each of them is composed of three convolutional
layers, with the first layers of the second and third stages performing
down-sampling. The decoder part also has three convolutional stages, with two
transpose convolutional layers for up-sampling.

At pyramid level $l$, given refined bi-directional flow $F_{0 \rightarrow 1}^l$
and $F_{1 \rightarrow 0}^l$, we can obtain the optical flow from input frames
$I_0^l$ and $I_1^l$ to target frame $I_t^l$ by linear scaling: $
F_{0 \rightarrow t}^l = t \cdot F_{0 \rightarrow 1}^l$, $F_{1 \rightarrow t}^l =
(1-t) \cdot F_{1 \rightarrow 0}^l$.  With $F_{0 \rightarrow t}^l$  and $F_{1
\rightarrow t}^l$, we forward-warp input frames $I_0^l$, $I_1^l$, and their
multi-scale context features $\{C_0^{l,0}, C_1^{l,0}\}$, $\{C_0^{l,1},
C_1^{l,1}\}$, $\{C_0^{l,2}, C_1^{l,2}\}$.  Furthermore, we generate an
initial estimate $\hat{I}_t^l$ of intermediate frame by up-sampling the
interpolation from previous $l+1$ level: $\hat{I}_t^l = \mathrm{up}_2(I_t^{l+1})
$. At top level, the initial estimate is set to the average of
two warped frames. Based on these, we feed the first encoder stage of the
synthesis module with warped frames, initial estimate of intermediate frame
$\hat{I}_t^l$, original input frames, and scaled bi-directional flow $F_{0
\rightarrow t}^l$ and $F_{1 \rightarrow t}^l$.  We feed warped context features
to the second and third encoder stages, and the first decoder stage to provide
multi-scale contextual cues.

The output of our frame synthesis module include two maps $M_0^l$ and $M_1^l$
for fusing two forward-warped frames $I_{0 \rightarrow t}^l$ and $I_{1
\rightarrow t}^l$, and a residual image $\Delta I_t^l$ for further refinement.
Then we can obtain refined intermediate frame $I_t^l$ by:
\begin{equation}
    I_t^l = \frac{(1-t) \cdot M_0^l \odot I_{0 \rightarrow t}^l + t \cdot M_1^l
        \odot I_{1 \rightarrow t}^l} {(1-t) \cdot M_0^l + t \cdot M_1^l}
    + \Delta I_t
    \label{eq:interp}
\end{equation}
where $\odot$ denotes element-wise multiplication.

Figure~\ref{fig:synthesis-module} illustrates our frame synthesis module. Its
design is inspired by previous context-aware synthesis
network~\cite{niklaus2018context,niklaus2020softmax,park2021asymmetric}, but has
two distinctive features. First, we feed the synthesis network with up-sampled
estimate of intermediate frame as an explicit reference for further refinement.
Second, our synthesis module is extremely lightweight, shared across pyramid
levels, and much simpler than the grid-like architecture in
~\cite{niklaus2018context,niklaus2020softmax,park2021asymmetric}.

\paragraph{Analysis of iterative synthesis for large
motion.}\label{sec:analysis-iter} Iterative synthesis from coarse-to-fine has
been proven beneficial for high-resolution images
synthesis~\cite{chen2017photographic}. In this work, we reveal that it can also
significantly improve the robustness of frame interpolation on large motion
cases.

To understand this, we start with a plain synthesis strategy, which does not
feed up-sampled estimate of intermediate frame into the frame synthesis module.
For plain synthesis, the actual synthesis is only performed on the bottom level,
because interpolations on previous levels have no path to affect
the frame synthesis on the bottom level. By comprehensively comparing the
interpolated frames by plain synthesis and iterative synthesis on large motion
cases, we draw empirical conclusions as follows.

\begin{figure}[tb]
\centering
\includegraphics[width=0.48\textwidth]{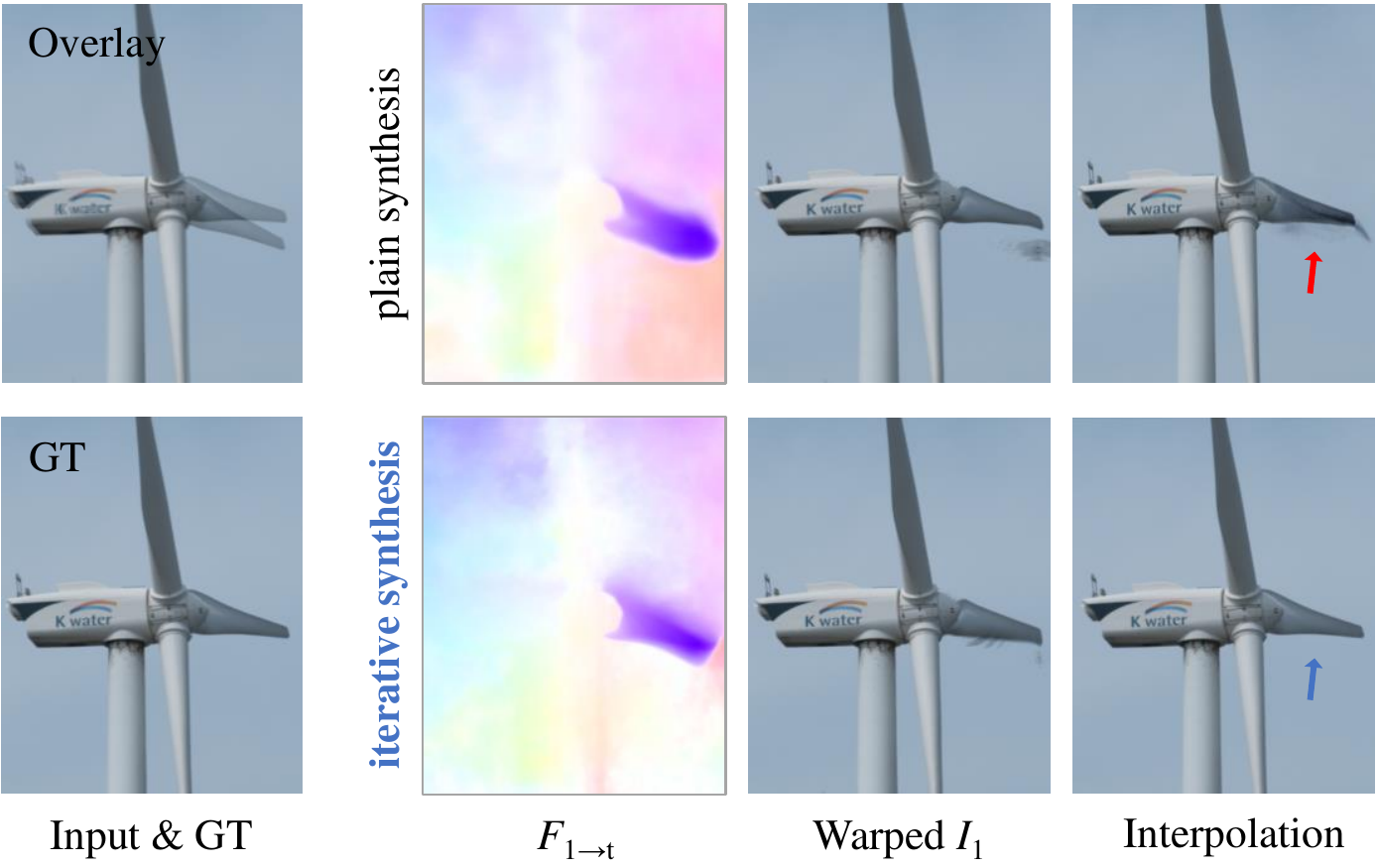}
\caption{Visualization of warped frames and interpolations for $t=1/8$, given a
    large motion case from 4K1000FPS~\cite{sim2021xvfi}. Our iterative synthesis
    enables robust interpolation even when warped frame has obvious artifacts.
    (Best viewed by zooming in.)
}
\vspace{-0.25cm}
\label{fig:comp-iter-syn}
\end{figure}

\begin{figure}[tb]
\centering
\includegraphics[width=0.48\textwidth]{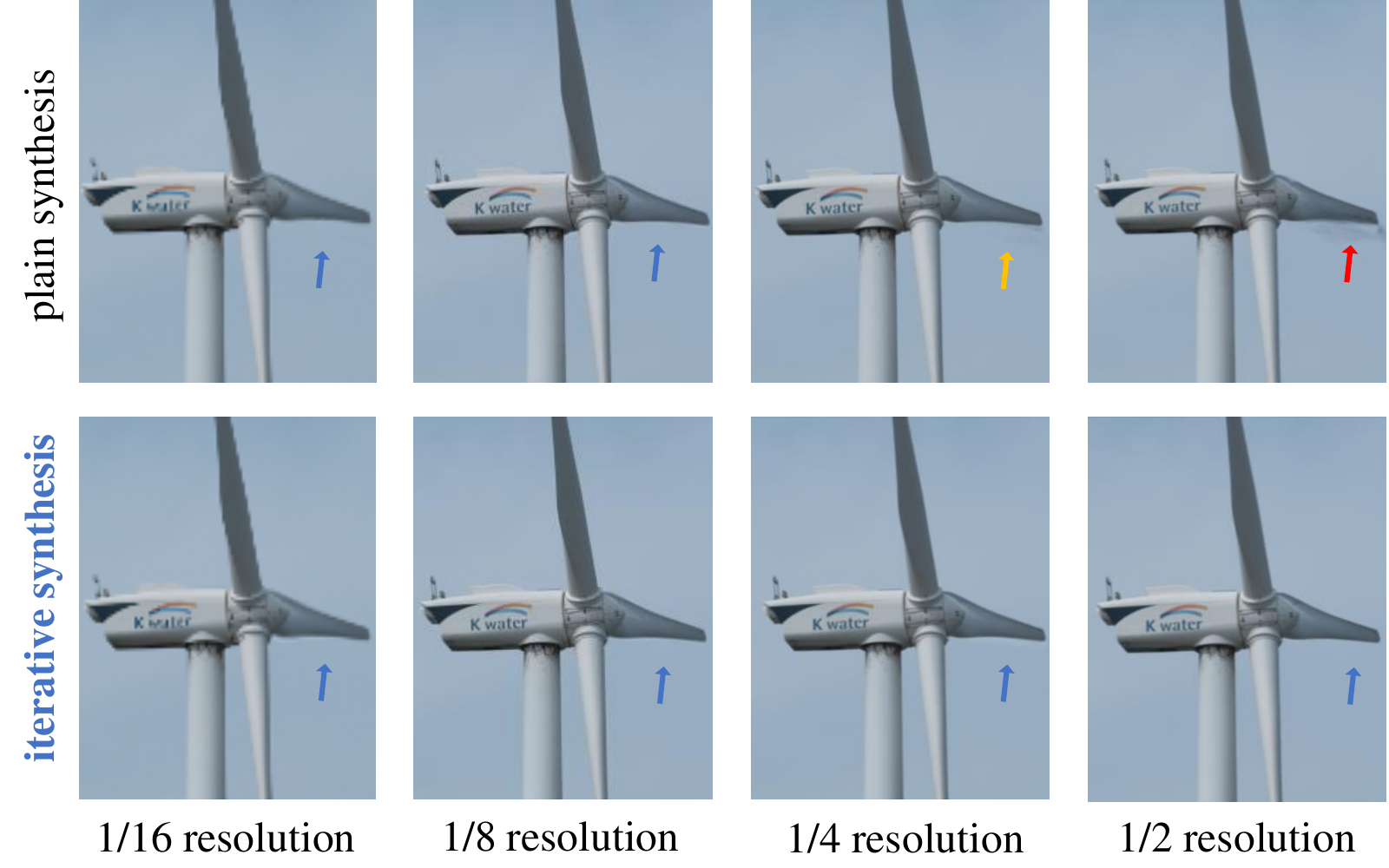}
\caption{Interpolated frames at different input resolutions, using plain and
    iterative synthesis. Images are intentionally resized to same size to show
    the differences. (Best viewed by zooming in.)
}
\vspace{-0.25cm}
\label{fig:progressive-iter-syn}
\end{figure}

(\romannumeral1) When using plain synthesis for interpolation, the obvious
artifacts due to large motion in forward-warped frames may lead to obvious
artifacts in interpolated frames. Figure~\ref{fig:comp-iter-syn} shows a typical
example.

(\romannumeral2) Coarse-to-fine iterative synthesis enables robust interpolation
even when the warped frame also has obvious artifacts (see
Figure~\ref{fig:comp-iter-syn}). We hypothesize that the up-sampled
interpolation, which is synthesized at lower-resolution pyramid level, may have
less or no artifacts due to smaller motion magnitude. Thus, it can guide the
synthesis module to produce robust interpolation at higher-resolution levels.

(\romannumeral3) Our hypothesis is supported by the evidence shown in
Figure~\ref{fig:progressive-iter-syn}, where we interpolate the same example on
reduced resolutions. We find that on 1/8 resolution, plain synthesis also gives
good interpolation without artifacts, since the motion magnitude is much
smaller. Our iterative synthesis gives good interpolation at all scales, because
it leverages the interpolations from low-resolution levels.

\subsection{Resolution-aware Adaptation in Testing}
The flexible design of UPR-Net allows resolution-aware adaptation during
inference. We describe two adaptation strategies for better estimating the
extremely large motions (in high-resolution videos) beyond the training phase.

\paragraph{Customizing the number of pyramid levels.}\label{para:pyr-lvls}
Benefiting from our recurrent module design, we can customize the number of
pyramid levels in testing to handle large motion. We follow the method proposed
in~\cite{jin2022enhanced}.  Assume that the number of pyramid levels in training
is $L^{train}$, and the width of test image is $n$ times of training images.
Then, the number of test pyramid levels is calculated as follows.

\begin{equation}
    L^{test} = \texttt{ceil}(L^{train} + \mathtt{log}_{2} n)
    \label{eq:level}
\end{equation}
where $\texttt{ceil}()$ rounds up a float number to an integer.

\paragraph{Skipping high-resolution levels for 4K input.} By default, at pyramid
level $l$, we up-sample optical flow and interpolation from level $l+1$ as
initializations.  But, actually, the design of our UPR-Net allows to up-sample
estimates from any previous levels if necessary.  We empirically verify that
reduced resolution is beneficial for accurate estimation of large motion for
extreme 4K resolution videos. We also verify that skipping last two
high-resolution levels for flow estimation and second-to-last level for frame
synthesis, can improve the accuracy of frame interpolation on 4K
benchmark~\cite{sim2021xvfi}.

\subsection{Architecture Variants}
We construct three versions of models by scaling the feature channels.
\begin{itemize}
    \item UPR-Net: Base version of UPR-Net. The numbers of channels of feature
        encoder's three stages are 16, 32, 64, respectively. The channel numbers
        of the 6-layer CNN in our optical flow module are 160, 128, 112, 96, 64,
        respectively. The channel numbers of the three encoder stages of our
        synthesis module are 32, 64, 128.
    \item UPR-Net large: A large version of UPR-Net, by scaling all feature
        channels of the base version by 1.5.
    \item UPR-Net LARGE: A larger version of UPR-Net, by scaling all feature
        channels of the base version by 2.0.
\end{itemize}

\begin{table*}[tb]
\centering
\small
\setlength{\tabcolsep}{0pt}
\begin{tabular*}{1.0\textwidth}{@{\extracolsep{\fill}}*{10}{lcccccccc}}
\hline
\multirow{2}{*}{methods} & \multirow{2}{*}{UCF101} &
\multirow{2}{*}{Vimeo90K} & \multicolumn{4}{c}{SNU-FILM} & parameters & runtime \\
\cline{4-7}
& &  &  easy & medium & hard & extreme & (millions) & (seconds)\\
\Xhline{2\arrayrulewidth}
DAIN~\cite{bao2019depth}  & 34.99/0.9683 & 34.71/0.9756 & 39.73/0.9902
                          & 35.46/0.9780 & 30.17/0.9335  & 25.09/0.8584 
                          & 24.0 & 0.151   \\
CAIN~\cite{choi2020channel}  & 34.91/0.9690 & 34.65/0.9730 & 39.89/0.9900 
                             & 35.61/0.9776  & 29.90/0.9292  & 24.78/0.8507 
                             & 42.8    & 0.037   \\
SoftSplat~\cite{niklaus2020softmax}  & 35.39/0.9520 & 36.10/0.9700
                                     & - & - & - & - & -  &  -  \\
AdaCoF~\cite{lee2020adacof} & 34.90/0.9680 & 34.47/0.9730 & 39.80/0.9900 
                            & 35.05/0.9754  & 29.46/0.9244  & 24.31/0.8439
                            & 22.9   & 0.030   \\
BMBC~\cite{park2020bmbc} & 35.15/0.9689 & 35.01/0.9764 & 39.90/0.9902
                         & 35.31/0.9774  & 29.33/0.9270  & 23.92/0.8432  
                         & 11.0 & 0.822  \\
CDFI full~\cite{ding2021cdfi} & 35.21/0.9500 & 35.17/0.9640 & 40.12/0.9906
                         & 35.51/0.9778  & 29.73/0.9277  & 24.53/0.8476  
                         & 5.0 & 0.172  \\
ABME~\cite{park2021asymmetric} & 35.38/\textcolor{blue}{\underline{0.9698}}
                               & 36.18/0.9805 & 39.59/0.9901 
                               & 35.77/0.9789 & 30.58/0.9364 & 25.42/0.8639
                               & 18.1 & 0.277   \\
XVFI$_v$~\cite{sim2021xvfi} & 35.18/0.9519 & 35.07/0.9681 & 39.78/0.9840 
                            & 35.37/0.9641
                            & 29.91/0.8935 & 24.73/0.7782  & 5.5   & 0.098   \\
RIFE~\cite{huang2020rife} & 35.28/0.9690 & 35.61/0.9780 & 40.06/0.9907
                          & 35.75/0.9789 & 30.10/0.9330 & 24.84/0.8534
                          & 9.8   &  \textcolor{red}{\textbf{0.012}}  \\

EBME-H*~\cite{jin2022enhanced} & 35.41/0.9697 & 36.19/0.9807 & 40.28/0.9910
                          & 36.07/0.9797 & 30.64/0.9368 & 25.40/0.8634
                          & 3.9   &  0.082  \\

VFIformer ~\cite{lu2022video} &\textcolor{blue}{\underline{35.43}}/\textcolor{red}{\textbf{0.9700}}
                  &\textcolor{red}{\textbf{36.50}}/\textcolor{red}{\textbf{0.9816}}
                                   & 40.13/0.9907 & 36.09/\textcolor{blue}{\underline{0.9799}}
                                   & 30.67/\textcolor{red}{\textbf{0.9378}}
                                   & 25.43/\textcolor{red}{\textbf{0.8643}} 
                                   & 24.1 & 1.018 \\
IFRNet ~\cite{kong2022ifrnet} & 35.29/0.9693 & 35.80/0.9794 & 40.03/0.9905
                                   & 35.94/0.9793 & 30.41/0.9358 & 25.05/0.8587
                                   & 5.0 & \textcolor{blue}{\underline{0.022}} \\
IFRNet large~\cite{kong2022ifrnet} 
                                   & 35.42/\textcolor{blue}{\underline{0.9698}}
                                   & 36.20/0.9808 & 40.10/0.9906
                                   & 36.12/0.9797
                                   & 30.63/0.9368
                                   & 25.27/0.8609 & 19.7 & 0.038 \\
\hline
UPR-Net  & 35.41/\textcolor{blue}{\underline{0.9698}} & 36.03/0.9801
         & 40.37/\textcolor{blue}{\underline{0.9910}} &36.16/0.9797 &30.67/0.9365 &25.49/0.8627
                  &\textcolor{red}{\textbf{1.7}}  & 0.042   \\
UPR-Net large  & \textcolor{blue}{\underline{35.43}}/\textcolor{red}{\textbf{0.9700}} 
               & 36.28/0.9810 & \textcolor{blue}{\underline{40.42}}/\textcolor{red}{\textbf{0.9911}}
                  &\textcolor{blue}{\underline{36.24}}/\textcolor{blue}{\underline{0.9799}}
                  &\textcolor{blue}{\underline{30.81}}/0.9370
                  &\textcolor{blue}{\underline{25.58}}/0.8636
                  &\textcolor{blue}{\underline{3.7}}  & 0.062  \\
UPR-Net LARGE  & \textcolor{red}{\textbf{35.47}}/\textcolor{red}{\textbf{0.9700}} 
                  &\textcolor{blue}{\underline{36.42}}/\textcolor{blue}{\underline{0.9815}}
                  &\textcolor{red}{\textbf{40.44}}/\textcolor{red}{\textbf{0.9911}}
                  &\textcolor{red}{\textbf{36.29}}/\textcolor{red}{\textbf{0.9801}}
                  &\textcolor{red}{\textbf{30.86}}/\textcolor{blue}{\underline{0.9377}}
                  &\textcolor{red}{\textbf{25.63}}/\textcolor{blue}{\underline{0.8641}}
                  &6.6  & 0.081  \\
\hline
\end{tabular*}
\caption{Qualitative (PSNR/SSIM) comparisons to state-of-the-art methods on
UCF101~\cite{soomro2012ucf101}, Vimeo90K~\cite{xue2019video} and
SNU-FILM~\cite{choi2020channel} benchmarks. \textcolor{red}{\textbf{RED}}: best
performance, \textcolor{blue}{\underline{BLUE}}: second best performance.}
\label{tab:sota}
\end{table*}

\section{Implementation Details}

\paragraph{Loss function.} 
Our loss is the sum of Charbonnier loss~\cite{charbonnier1994two} and
census loss~\cite{meister2018unflow} between ground truth $I_t^{GT}$ and our
interpolation $I_{t}$ estimated at the bottom pyramid level:
\begin{equation}
    L = \rho (I_t^{GT} - I_t) + L_{cen}(I_t^{GT}, I_t).
    \label{eq:loss}
\end{equation}

% where $\rho(x) = (x^2 + \epsilon^2)^\alpha$ is the Charbonnier function,
% $L_{cen}$ is the census loss. We empirically set $\alpha=0.5$,
% $\epsilon=10^{-6}$.

\paragraph{Training dataset.} The Vimeo90K dataset~\cite{xue2019video} contains
51,312 triplets with resolution of $448\times256$ for training. We augment the
training images by randomly cropping $256\times256$ patches. We also apply
random flipping, rotating, reversing the order of the triplets for augmentation.

\paragraph{Pyramid levels in training.} We use 3-level image pyramids for our
UPR-Net during training, which is sufficient to capture the motions on
Vimeo90K~\cite{xue2019video}. 

\paragraph{Optimization.} We use the AdamW~\cite{loshchilov2017decoupled}
optimizer with weight decay $10^{-4}$ for 0.8 M iterations, and batch size
of 32. We gradually reduce the learning rate during training from
$2\times10^{-4}$ to $2\times10^{-5}$ using cosine annealing.

\section{Experiments}
\label{sec:exp}

\subsection{Experiment Settings}

\paragraph{Evaluation datasets.} While our models are trained only on
Vimeo90K~\cite{xue2019video}, we evaluate them on a broad range of benchmarks with
different resolutions.
\begin{itemize}
    \item \textbf{UCF101~\cite{soomro2012ucf101}}: The test set of UCF101
        contains 379 triplets with a resolution of 256$\times$256. UCF101
        contains a large variety of human actions.
    \item \textbf{Vimeo90K~\cite{xue2019video}}: The test set of Vimeo90K
        contains 3,782 triplets with a resolution of 448$\times$256.
    \item \textbf{SNU-FILM~\cite{choi2020channel}}: This dataset contains 1,240
        triplets, and most of them are of the resolution around 1280$\times$720.
        It contains four subsets with increasing motion scales -- easy, medium,
        hard, and extreme.
    \item \textbf{4K1000FPS~\cite{sim2021xvfi}}: This is a 4K resolution
        benchmark (X-TEST) that enables multi-frame ($\times$8) interpolation.
\end{itemize}

\paragraph{Resolution-aware adaptation.} According to Equation~\ref{eq:level},
we set the test pyramid levels for UCF-101, SNU-FILM and 4K1000FPS as 3, 5 and
7, respectively. We skip the last two pyramid levels for
bi-directional flow and second-to-last level for frame synthesis on 4K1000FPS.
We also report the results without level skipping on 4K1000FPS.

\paragraph{Metrics.} PSNR and SSIM~\cite{wang2004image} are used for
quantitative evaluation of frame interpolation. For running time, we test all
models with a RTX 2080 Ti GPU under $640\times480$ resolution, and average the
running time by 100 iterations.

\begin{figure*}[!htb]
\centering
\includegraphics[width=1.0\textwidth]{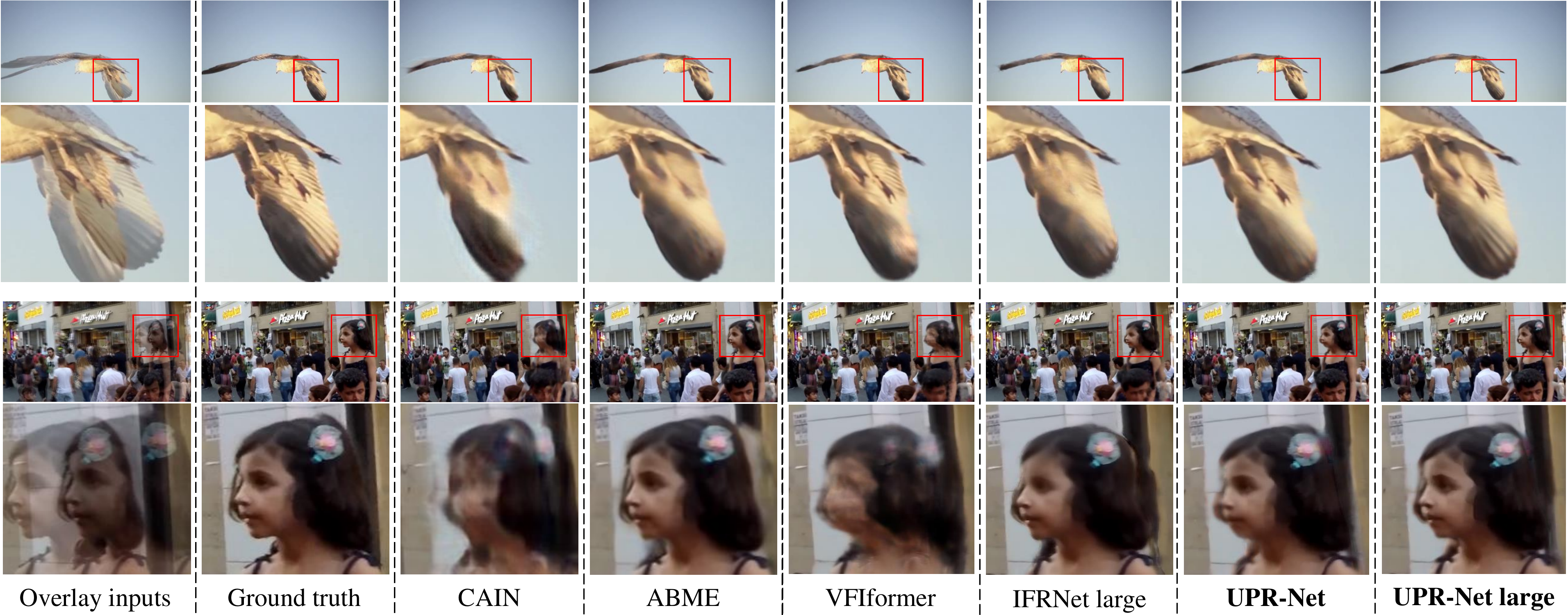}
\caption{
    Qualitative comparisons on SNU-FILM~\cite{choi2020channel}. First example
    from the hard subset, and second example from extreme subset.
}
\label{fig:vis-snufilm}
\end{figure*}

\begin{figure*}[!htb]
\centering
\includegraphics[width=1.0\textwidth]{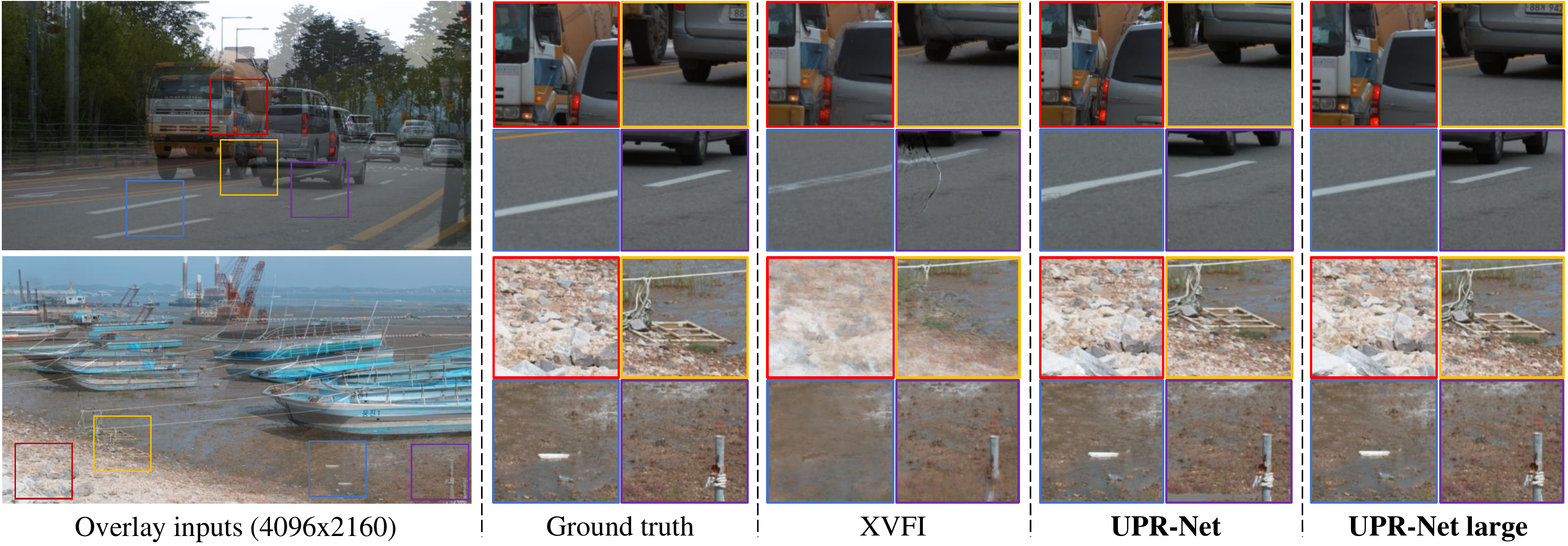}
\caption{
    Qualitative comparisons on X-TEST of 4K1000FPS~\cite{sim2021xvfi}. First
    example interpolated at $t=1/8$ and second example at $t=1/2$.
}
\vspace{-0.25cm}
\label{fig:vis-4k1000fps}
\end{figure*}

\subsection{Comparisons with State-of-the-art Methods}

We compare with state-of-the-art methods, including DAIN~\cite{bao2019depth},
CAIN~\cite{choi2020channel}, SoftSplat~\cite{niklaus2020softmax},
AdaCoF~\cite{lee2020adacof}, BMBC~\cite{park2020bmbc}, CDFI~\cite{ding2021cdfi},
ABME~\cite{park2021asymmetric},  XVFI~\cite{sim2021xvfi},
RIFE~\cite{huang2020rife}, EBME~\cite{jin2022enhanced},
VFIformer~\cite{lu2022video}, and IFRNet~\cite{kong2022ifrnet}. We report
results by executing the source code and trained models, except for SoftSplat
which has not released the full code. For SoftSplat, we copy the results from
original paper.

\paragraph{Low resolution frame interpolation.}
Table~\ref{tab:sota} reports the comparison results on low-resolution UCF101 and
Vimeo90K datasets. The transformer-based VFIformer~\cite{lu2022video} achieves
the best accuracy on Vimeo90K, which however has a large number of parameters,
and runs very slow.  Our UPR-Net LARGE model achieves the best performance on
UCF101 and second best result on Vimeo90K.  Furthermore, our UPR-Net large and
UPR-Net models also achieve excellent accuracy on these two benchmarks. In
particular, our UPR-Net model, which only has 1.7M parameters, outperforms many
recent large models on UCF101, including SoftSplat~\cite{niklaus2020softmax} and
ABME~\cite{park2021asymmetric}.

\paragraph{Moderate resolution frame interpolation.}
Table~\ref{tab:sota} also reports the comparison results on SNU-FILM.  When
measured with PSNR, our UPR-Net series outperform all previous state-of-the-art
methods. In particular, our base UPR-Net model outperforms the large VFIformer
model, in part due to its capability of handling challenging motion.

Figure~\ref{fig:vis-snufilm} gives two examples from the hard  and extreme
subsets from SNU-FILM, respectively. Our methods produce better interpolation
results than IFRNet large model~\cite{kong2022ifrnet} for local textures (first
two rows), and give promising results for large motion cases (last two rows),
much better than CAIN~\cite{choi2020channel} and VFIformer~\cite{lu2022video},
and sightly better than ABME~\cite{park2021asymmetric} and IFRNet
large~\cite{kong2022ifrnet}.

\paragraph{4K resolution multiple frame interpolation.}

\begin{table}[tb]
\centering
\small
\setlength{\tabcolsep}{0pt}
\begin{tabular*}{0.48\textwidth}{@{\extracolsep{\fill}}*{5}{lcccc}}
\hline
methods & arbitrary time & reuse
flow & X-TEST (4K) \\
\Xhline{2\arrayrulewidth}
DAIN~\cite{bao2019depth}  & \checkmark & \checkmark &  26.78/0.8065   \\
AdaCoF~\cite{lee2020adacof} & $\times$ & $\times$  & 23.90/0.7271  \\
ABME~\cite{park2021asymmetric} & \checkmark & $\times$  
                               & 30.16/0.8793 \\
XVFI~\cite{sim2021xvfi} & \checkmark & partial & 30.12/0.8704  \\
% IFRNet large~\cite{kong2022ifrnet} & \checkmark & $\times$ & 24.84/0.8012  \\
\hline
UPR-Net & \checkmark & \checkmark & 30.13/0.8990  \\
UPR-Net large & \checkmark & \checkmark 
                    & \textcolor{red}{\textbf{30.68}}/\textcolor{red}{\textbf{0.9086}}  \\
UPR-Net LARGE & \checkmark & \checkmark 
                    &
                    \textcolor{blue}{\underline{30.50}}/\textcolor{blue}{\underline{0.9048}}  \\
\hline
UPR-Net$\dag$ & \checkmark & \checkmark & 29.27/0.8877  \\
UPR-Net large$\dag$ & \checkmark & \checkmark 
              & 30.14/0.9046  \\
UPR-Net LARGE$\dag$ & \checkmark & \checkmark 
              & 29.91/0.8998  \\
\hline
\end{tabular*}
\caption{Quantitative (PSNR/SSIM) comparisons on X-TEST (from
4K1000FPS~\cite{xue2019video}) for 8x multi-frame interpolation. $\dag$: not
skip pyramid levels for optical flow and frame synthesis.}
\label{tab:multi-interp} \end{table}

Table~\ref{tab:multi-interp} reports the 8x interpolation results on X-TEST.
Our UPR-Net large model achieves the best performance. Skipping proper
high-resolution levels enables better interpolation results, as it benefits the
estimation of extremely large motions. Our UPR-Net LARGE model does not achieve the
best results, as it might overfit the motion magnitude of
Vimeo90K~\cite{xue2019video}.  Furthermore, our method enables arbitrary-time
frame interpolation, and the bi-directional flow only needs to be estimated once
for multi-frame interpolation.

Figure~\ref{fig:vis-4k1000fps} shows two interpolation examples. Our models are
robust to large motion, and give better interpolation for local textures than
XVFI~\cite{sim2021xvfi}.

\paragraph{Parameter and inference efficiency.} As shown the last two columns in
Table~\ref{tab:sota}, our base version of UPR-Net only has 1.7 M parameters.
UPR-Net is much faster than recent ABME and VFIformer, but slower than RIFE and
IFRNet.

\subsection{Ablation Studies of Design Choices}

In Table~\ref{tab:ablation}, we present ablation studies of the design choices
of our UPR-Net on Vimeo90K~\cite{xue2019video}, the hard subset of
SNU-FILM~\cite{choi2020channel}, and X-TEST of 4K1000FPS~\cite{sim2021xvfi}.

\paragraph{Recurrent module design.} We verify the effectiveness of recurrent
model design by replacing the optical flow module or frame synthesis module with
three cascaded non-recurrent modules.  These two non-recurrent counterparts can
still achieve promising performance on low-resolution
Vimeo90K~\cite{xue2019video}.  However, they can not customize the number of
pyramid levels in testing, and thus lack of flexibility in handling large
motion. As a result, they can not achieve good results on the hard subset of
SUN-FILM~\cite{choi2020channel}, and the extreme 4K-resolution X-TEST
benchmark~\cite{sim2021xvfi}.

\paragraph{Iterative frame synthesis.} Plain synthesis without using up-sampled
interpolation achieves good performance on Vimeo90K~\cite{xue2019video}.
This suggests that iterative synthesis might be unnecessary for low-resolution
images.  Compared to many state-of-the-art methods, it also achieves good
results on hard subset of SNU-FILM~\cite{choi2020channel}. Even so, iterative
synthesis achieves better performance on hard subset, and much better
performance on X-TEST~\cite{sim2021xvfi}.  These results provide quantitative
evidences about our analysis in Section~\ref{sec:analysis-iter}.

\begin{table}[tb]
\centering
\small
\setlength{\tabcolsep}{0pt}
\begin{tabular*}{0.48\textwidth}{@{\extracolsep{\fill}}*{8}{c}}
\hline
experiments & methods & Vimeo90K & Hard & X-TEST & param.\\
\Xhline{2\arrayrulewidth}
\multirow{3}{*}{recurrent modules}
            & only flow & 35.98 & 30.33 & 24.01 & 3.60M \\
                              & only syn. & \textbf{36.03} & 30.25 & 23.65
                              & 2.68M\\
                             & \cellcolor{gray!40}both & \textbf{36.03}
                             & \textbf{30.67} & \textbf{30.13} & 1.65M\\
\hline
\multirow{2}{*}{iterative synthesis}
                             & plain & 36.02 & 30.57 & 28.91 & 1.65M \\
                             & \cellcolor{gray!40}iterative  & \textbf{36.03}
                             & \textbf{30.67} & \textbf{30.13} & 1.65M\\
\hline
\multirow{2}{*}{unified pipeline}
                             & separate & 35.96 & \textbf{30.67} & 29.89 & 1.82M\\
                             & \cellcolor{gray!40}unified & \textbf{36.03}
                             & \textbf{30.67} & \textbf{30.13} & 1.65M\\
\hline

% \multirow{2}{*}{smaller flow module}
%                              & plain syn. & 35.81 & 30.46  & 27.61 \\
%                              & \cellcolor{gray!40} iterative syn.
%                              & \textbf{35.88} & \textbf{30.62}  & \textbf{29.66} \\
% \hline

% \multirow{2}{*}{warp approxiation}
%                              & plain syn. & \textbf{31.09} & \textbf{28.67}  &
%                              \textbf{26.33} \\
%                              & \cellcolor{gray!40} iterative syn. & 31.02 & 28.40  & 26.05 \\
% \hline
\multirow{2}{*}{correlation volume}  
                               & without & 35.81 & 30.59 & 29.51 & 1.64M \\
                             & \cellcolor{gray!40}with & \textbf{36.03} & \textbf{30.67}
                             & \textbf{30.13} & 1.65M\\
\hline
\multirow{2}{*}{context feature}  
                             & without & 35.63 & \textbf{30.68} & 29.91 & 1.43M\\
                              & \cellcolor{gray!40}with & \textbf{36.03} & 30.67
                              & \textbf{30.13} & 1.65M \\
\hline
% \multirow{3}{*}{training loss}  
%                               & multi-scale & 35.58 & 30.53 & 30.03 \\
%                               & \cellcolor{gray!40}single-scale & \textbf{36.03}
%                              & \textbf{30.67} & 30.13 \\
%                               & w/o census & 35.98 & 30.59 & \textbf{30.32} \\
% \hline
% \multirow{5}{*}{test pyramid levels} & 3 levels  & \textbf{36.03} & 30.30 
%                                      & 23.59 \\
%                                   & 4 levels  & 35.92 
%                                   & \textbf{30.69} & 25.97 \\
%                                   & 5 levels  & 35.89 & 30.67 & 29.16\\
%                                   & 6 levels & 35.83 & 30.64 & \textbf{30.15} \\
%                                   & 7 levels & 35.83 & 30.64 & 30.13\\
% \hline
% \multirow{4}{*}{skip high-res. levels}  & skip 0 level & \textbf{36.03} 
%                                       & 30.67  & 29.27\\
%                                   & skip 1 level  & 35.41 
%                                   & \textbf{30.69} & 29.60\\
%                                   &  skip 2 levels &33.79 
%                                   & 30.41 & \textbf{30.13}\\
%                                   &  skip 3 levels & 27.04 & 28.70 & 29.42\\
% \hline
% \end{tabularx}
\end{tabular*}
\caption{Ablation studies of our design choices on Vimeo90K~\cite{xue2019video},
hard subset of SNU-FILM~\cite{choi2020channel}, and 4K
X-TEST~\cite{sim2021xvfi}. Default settings (independent of benchmark datasets)
are marked in \colorbox{gray!40}{gray}.
}
\label{tab:ablation}
\end{table}

\paragraph{Unified pipeline.} We construct a pipeline consisting of two separate
pyramid recurrent networks (with separate feature encoders) for bi-directional
flow and frame synthesis.  Like previous
works~\cite{sim2021xvfi,jin2022enhanced}, optical flow network outputs
single-scale bi-directional flow from the bottom pyramid level.  We down-sample
the bi-directional flow to generate multi-scale flow for iterative frame
synthesis. This separate counterpart adds about 14\% parameters (and
computational cost), and achieves slight inferior performance to our unified
pipeline. We prefer our unified pipeline, due to its simplicity and elegance.
But, it is worth noting that our pyramid recurrent frame synthesis
network can be combined with any existing optical flow model for frame
synthesis.

% \paragraph{Smaller optical flow module.} We construct a smaller optical flow
% module by reducing the channel number to half, and keep other modules unchanged.
% The performance degradation of plain synthesis is more severe than that of
% iterative synthesis, suggesting that plain synthesis is more sensitive to the
% capacity of optical flow module.

\paragraph{Correlation volume for bi-directional flow.} Removing the correlation
volume from our optical flow module leads to performance degradation on all
benchmarks. This result is consistent with the observations
in~\cite{jin2022enhanced}. Arguably, the importance of correlation volume is
overlooked by many existing frame interpolation
methods~\cite{zhang2020flexible,sim2021xvfi,huang2020rife}.

\paragraph{Context feature for frame synthesis.} Removing the context features
from our synthesis module will greatly degrade the accuracy on
Vimeo90K~\cite{xue2019video}. But, surprisingly, it does not lead to obvious
inferior performance on large motion benchmarks, which is worth further
investigation.

% \paragraph{Training loss.} TODO

% \paragraph{Number of pyramid levels in testing.} TODO

% \paragraph{Skipping high-resolution levels in testing.} TODO

\section{Conclusion and Future Work}
\label{sec:conclusion}

This work presented UPR-Net, a lightweight Unified Pyramid Recurrent Network for
frame interpolation. UPR-Net achieved excellent performance on various frame
interpolation benchmarks. In the future, we will investigate some
interesting problems related to this work. Firstly, we will verify the
generalization of our iterative synthesis by replacing our motion estimator with
an off-the-shelf optical flow model (\eg, PWC-Net~\cite{sun2018pwc}). Second, we
will train our models with the Vimeo90K-Septuplet~\cite{xue2019video} to
investigate whether multi-frame interpolation during training is beneficial for
multi-frame interpolation in testing.

%%%%%%%%% REFERENCES
{\small
\bibliographystyle{ieee_fullname}
\bibliography{upr}
}

\end{document}